\pdfoutput=1

\documentclass[11pt]{article}

\usepackage[preprint]{acl}

\usepackage{times}
\usepackage{latexsym}

\usepackage[T1]{fontenc}

\usepackage[utf8]{inputenc}

\usepackage{microtype}

\usepackage{inconsolata}

\usepackage{graphicx}

\usepackage{algorithm}
\usepackage{algorithmic}
\usepackage{amsmath}
\usepackage{mathtools}
\usepackage{amsthm}
\usepackage{subcaption}
\usepackage{booktabs}

\newcommand{\loss}{\mathsf{loss}}
\DeclareMathOperator*{\argmax}{arg\,max}

\newtheorem*{remark}{Remark}

\title{An Efficient Rehearsal Scheme for Catastrophic Forgetting Mitigation during Multi-stage Fine-tuning}

\author{\textbf{Andrew Bai\textsuperscript{1}} \thanks{Work done while the author was a student researcher at Google Cloud AI Research. Correspondence to: Andrew Bai
 <\texttt{andrewbai@cs.ucla.edu}>, Ankur Taly <\texttt{ataly@google.com}>}, \textbf{Chih-Kuan Yeh\textsuperscript{2}}, \textbf{Cho-Jui Hsieh\textsuperscript{1,2}}, \textbf{Ankur Taly\textsuperscript{2}} \\
  \textsuperscript{1}University of California, Los Angeles, \textsuperscript{2}Google Inc.
  }
  
\begin{document}
\maketitle
\begin{abstract}
Incrementally fine-tuning foundational models on new tasks or domains is now the de facto approach in NLP.
A known pitfall of this approach is the \emph{catastrophic forgetting} of prior knowledge that happens during fine-tuning.
A common approach to alleviate such forgetting is to rehearse samples from prior tasks during fine-tuning.
Several existing works assume a fixed memory buffer to store prior task examples, while relying on inferences (forward passes) with the model at hand for choosing examples for rehearsal from the buffer.
However, given the increasing computational cost of model inference, and decreasing cost of data storage, we focus on the setting to rehearse samples with a fixed computational budget instead of a fixed memory budget.
We propose a sampling scheme, \texttt{\bf mix-cd}, that prioritizes rehearsal of ``collateral damage'' samples, which are samples predicted correctly by the prior model but forgotten by the incrementally tuned one.
The crux of our scheme is a procedure to efficiently estimate the density of collateral damage samples without incurring additional model inferences.
Our approach is computationally efficient, easy to implement, and outperforms several leading continual learning methods in compute-constrained settings.
All the code will be publicly available at \url{https://github.com/jybai/mix-cd-rehearsal}.

\end{abstract}

\section{Introduction}
The advent of pretrained foundational models has led to a paradigm shift in machine learning, wherein, a single model can be trained to learn a wide variety of tasks.
Incrementally learning of a new task or domain is carried out by fine-tuning some or all parameters on the new task.
Such learning is both compute and data-efficient as it benefits from the patterns learned during learning of previous tasks (as well as pretraining).
It is common to sequentially fine-tune foundational models over various datasets to teach the model new tasks or improve performance on new domains for an already learned task.

\begin{figure*}[ht]
\small
    \centering
    \includegraphics[width=\textwidth]{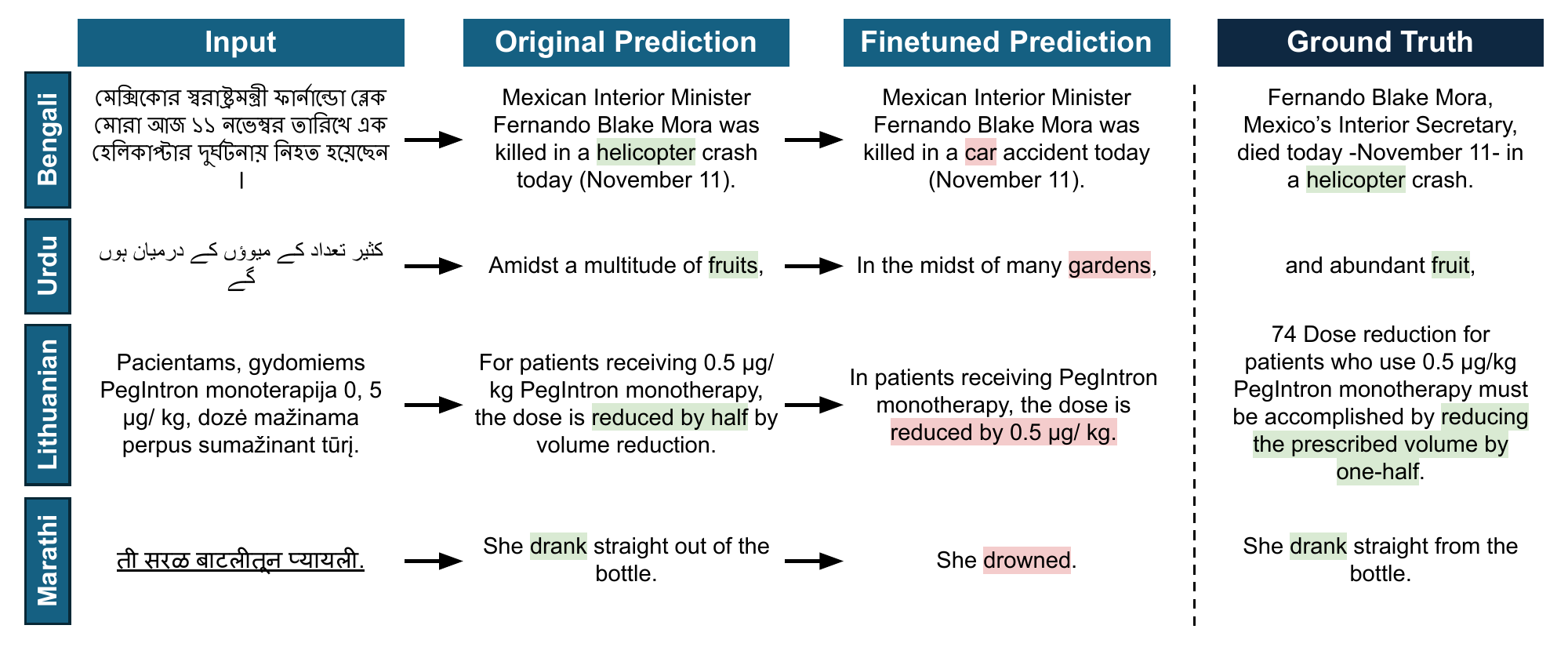}
    \vspace*{-5mm}
    \caption{Examples of collateral damage in prior language translations after fine-tuning on Danish-to-English.}
    \label{fig:nmt-example}
\end{figure*}

Unfortunately, such incremental tuning of the parameters may lead to forgetting of tasks or domains learned previously.
For instance, consider a multilingual translation model that can translate from other languages to English.
When we incrementally tune this model to learn translation from an additional language (e.g. Danish), we find that its performance degrades on previously learned languages; see Figure~\ref{fig:nmt-example}.
Similar forgetting of prior skills and knowledge happens when instruction-tuned language models are aligned on human preferences using reinforcement learning; this is referred to as the \emph{alignment tax}~\cite{lin2024mitigating}.

In this work, we study computationally efficient methods for incrementally training foundational models on new tasks or domains, while preventing such \emph{catastrophic forgetting} of knowledge from selected previous tasks. 
A common strategy to reduce catastrophic forgetting during fine-tuning is to ``rehearse'' samples from previous tasks by mixing them into the fine-tuning set.
The rehearsal samples are typically drawn from a limited rehearsal buffer holding samples from previous tasks.


However, there are two main criticisms of existing rehearsal settings. 
First, most rehearsal methods assume only a small rehearsal buffer, citing storage costs and data access restrictions as the reason.
This limits the space for drawing rehearsal samples, which can lead to overfitting~\cite{verwimp2021rehearsal}.
Second, many rehearsal methods require high computational costs, in the form of inferencing with the model at hand, to select examples for rehearsal. 
Many existing rehearsal methods fall short when we take into account the computational cost of sampling examples for rehearsal.
Recent work~\cite{prabhu2023computationally} shows that several high-performing methods are unable to beat random uniform rehearsal in compute-constrained settings.

Over the last decade, storage costs have dramatically reduced to nearly 2 cents/gb~\cite{prabhu2023computationally}.
On the other hand, the size of foundational models has grown exponentially, keeping computational costs\footnote{Performing inference on N tokens with a transformer model with D parameters requires approximately 2ND FLOPs. Thus, inferencing on a sequence of 100 tokens with a 1B parameter model would involve a staggering $2\cdot10^{11}$ FLOPS.}
of training and inference high. 
Thus, in this work, we seek rehearsal methods that are computationally efficient but are allowed full access to prior fine-tuning sets.
We assume a setting where the multi-stage fine-tuning is performed by the same party, and therefore there are no data access restrictions.


In this work, we propose \textbf{mix-cd}, a rehearsal method that is no more expensive than random uniform rehearsal but achieves a strictly better tradeoff between new and previous task performances.
This is significant as uniform sampling had been demonstrated to be a strong baseline in compute-constrained settings~\citep{prabhu2020greedy}.
The key insight in our method is that it is beneficial to prioritize rehearsing \emph{collateral damage} samples.
Collateral damage is defined as being predicted correctly by the existing model, but incorrectly by the incrementally tuned one.

A key technical challenge is that the naive approach for obtaining collateral damage information requires making a forward pass with the fine-tuned model on the prior dataset.
This incurs significant computation costs.
To overcome this, we propose an efficient method for estimating the collateral damage density within the data distribution.
The estimated density is updated throughout the fine-tuning process to keep track of the dynamic changes where collateral damage occurs.

Overall, our scheme retains the desirable quality of being general, lightweight, and easy to implement, and can serve as a drop-in replacement for the random uniform rehearsal approach.
Through experiments on multiple tasks, we demonstrate that our scheme outperforms random uniform rehearsal and several other offline and online continual learning baselines in striking a favorable trade-off between new and previous task performances.

\section{Background and Related Work}

\subsection{Multi-stage fine-tuning framework}
The multi-stage fine-tuning framework finds applications across various domains and tasks within machine learning.
In natural language processing, pretrained language models such as BERT~\citep{devlin2018bert}, T5~\citep{raffel2020exploring}, and others are extensively fine-tuned for specific tasks like sentiment analysis~\citep{sun2019fine}, text summarization~\citep{liu2019text}, and question answering~\citep{roberts2020much}. 
Large generative language models such as GPT~\citep{brown2020language} and Llama~\citep{touvron2023llama} are instruction-tuned~\citep{wei2021finetuned} on human-provided feedback to align their generation with human responses.
In computer vision, pretrained vision transformers are commonly fine-tuned for image classification, object detection~\citep{li2022exploring}, and segmentation tasks~\citep{thisanke2023semantic}. 
Transfer learning through continual fine-tuning is also prevalent in medical imaging~\citep{he2023transformers} for tasks like disease diagnosis and organ segmentation. 

\subsection{Retaining Prior Performance}
One major challenge for multi-stage fine-tuning is retaining prior performance while improving on the current fine-tuning task.
In some cases where the fine-tuned model is only expected to perform well on a limited set of fine-tuned examples, in which case, disregarding the prior task is acceptable.
On the other hand, studies have shown that maintaining the prior performance is beneficial to not overfit the fine-tuning data and other desiderata~\citep{lin2023speciality,he2021analyzing}.


\paragraph{Forgetting prevention by weight regularization}
Weight regularization~\citep{lin2023speciality} methods prevent prior task forgetting by directly restricting the weights of the fine-tuned model.
The weights can be constrained during fine-tuning by anchoring them to the prior model weights~\citep{panigrahi2023task,xuhong2018explicit,kirkpatrick2017overcoming}.
The constraint can also be in the form of low-rank weight adaptation with LoRA~\citep{hu2021lora}.
On the other hand, \citet{wortsman2022robust} proposes WiSE-FT to ensemble the prior and fine-tuned weights post-hoc to achieve a balanced tradeoff of performance between tasks.
In general, weight regularization methods rely on the assumption that the new model optima post-fine-tuning lies close to the prior optima.



\paragraph{Forgetting prevention by rehearsal}
Rehearsal-based methods prevent prior task forgetting by including a portion of prior data in the fine-tuning phase.
A common approach is to sample uniformly at random from the prior data and mix them into the fine-tuning set~\citep{he2021analyzing,kazemi2023understanding}.
Some prior works consider the setting where prior data must be selected offline before accessing the next task.
\citet{yoon2022online} proposed Online Coreset Selection, which selects important samples while streaming through the prior dataset.
They prioritize data points with high minibatch similarity and sample diversity.
\citet{mok-etal-2023-large} proposed Dynamic Instance Selection, which selects the highest and lowest predictive entropy samples to allow easier and more difficult examples to be represented evenly.
However, such offline selection methods fail to consider the impact of the new fine-tuning task and are unable to tailor the selected samples to best mitigate the induced forgetting.
\citet{aljundi2019online} proposed Maximally Interfered Sampling (MIR), where high loss difference points are sampled from a small replay buffer.~\citet{prabhu2023computationally} has shown that all existing continual learning methods evaluated fail to beat the random mixing baseline in a computationally-constrained setting without the memory constraint. Our work adopts the computationally-constrained setting motivated by \citet{prabhu2023computationally}.

\section{Evaluation Protocol and Key Idea}
Our objective is to: \emph{Improve performance on the fine-tuned task while avoiding performance deterioration on prior tasks.} 
In this section, we define our evaluation protocol and motivate the design of our method via some key empirical observations.

\subsection{Evaluation Protocol}
We start with a model trained on a prior task and assume that the training losses on the prior task examples are stored and accessible without any extra computational costs. 
Our objective is to improve the fine-tuned task performance while balancing prior task performance given a limited computational budget. 
Thus, we compare different rehearsal strategies by examining the Pareto curve of the prior (y-axis) and fine-tuned (x-axis) task performances.

Different points on the same Pareto curve correspond to different instantiations of the same rehearsal strategy with different \emph{mix ratios} $\beta$ given a fixed computational budget $c$.
Mix ratio is defined as the proportion of the computation budget allocated to rehearsing the prior task and fine-tuning on the new task.
For example, if $\beta=0.1$ then $c_p := 0.1* c$ is the budget allocated to rehearsal, and $c_f := 0.9*c$ to fine-tuning.
The rehearsal budget includes both the costs of sampling and training on the rehearsal instances.
The fine-tuning budget includes the cost of training on the new task instances.
By ablating the mix ratios, the Pareto curves are formed with the computational budget being traded between rehearsing and fine-tuning.
Every point on the Pareto frontier shares the same computational cost and differs in how the computational budget is allocated between the prior and fine-tuned tasks.
Methods with a Pareto frontier towards the top right direction are preferable.

\begin{figure}[ht]
    \centering
    \includegraphics[width=\columnwidth]{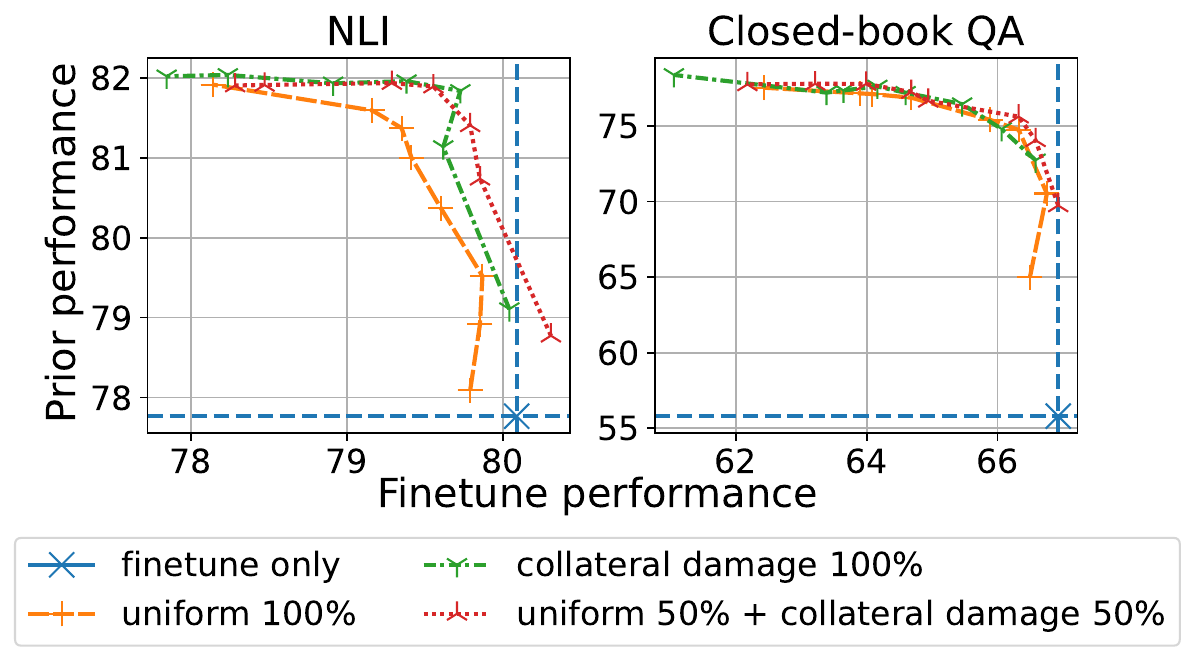}
    \caption{Preliminary observations suggest that while random rehearsal of prior data helps mitigate collateral damage, upweighting collateral damage samples in the prior data distribution benefits the joint performance on both tasks even more. 
    }
    \label{fig:pf-prelim}
\end{figure}

\subsection{Formal Definition of Collateral Damage}
Let us denote the base (prior) model as $f$, the fine-tuned model as $f'$, prior data samples as $z_p$, and fine-tuned data samples as $z_f$.
Let $\phi$ denote the indicator function for collateral damage.
For classification tasks, a sample $z_p = (x,y)$ suffers from collateral damage, denoted by $\phi_{f, f'}(z_p)$, if it is predicted correctly by $f$ but incorrectly by $f'$.

\vspace{-4mm}
\begin{align*}
    \phi_{f, f'}(z_p) := &\big(\argmax f(x) \equiv y\big) \wedge \\
    &\big(\argmax f'(x) \neq y \big)
\end{align*}

For non-classification tasks, collateral damage can be defined using the losses of the base and fine-tuned models.
Specifically, a sample $z_p$ suffers from collateral damage if its loss on $f$ is less than a threshold $\tau$, and loss on $f'$ is greater than $\tau$.
\[
    \phi_{f, f'}^{\tau}(z_p) = \big(\loss(f, z_p) < \tau \big) \wedge \big(\loss(f', z_p) > \tau \big)
\]
In our experiments, we set $\tau$ as the $90^\text{th}$ percentile of the loss of the prior model on the prior data.

\subsection{Key Idea: Rehearse Collateral Damage}
\label{sec:key_insight_cd}
Our key idea is to sample and train on more \emph{collateral damage} data during rehearsal, i.e., examples that were predicted correctly by the prior model but were ``forgotten'' during fine-tuning. 

\paragraph{Uniform sampling disregards the utility of samples.}
We motivate prioritizing collateral damage samples by ablating random uniform rehearsal mixed with different proportions of collateral damage samples in Fig~\ref{fig:pf-prelim}.
The fine-tuned baseline (with no rehearsal) suffers significant collateral damage as the prior task is catastrophically forgotten.
Random uniform rehearsal helps retain the prior task performance at the cost of worse fine-tuned performance.
However, random uniform rehearsal is sub-optimal as it does not account for the ``utility'' of the prior samples.
Mixing in $50\%$ collateral damage samples improves the Pareto curve and achieves better joint performance in both experiment settings. 
We hypothesize samples predicted correctly before fine-tuning but incorrectly after, are more easily to be predicted correctly again after rehearsal.


\paragraph{High computation cost for acquiring collateral damage signal.}
One major technical challenge is that determining whether a sample is collateral damage requires at least an additional inference on the current fine-tuned model.
This makes the cost of sampling collateral damage samples significantly higher than random uniform rehearsal, which has a negligible sampling cost.
Recall the computational budget for rehearsal $c_p$ is split into budgets for sampling $c_{p,s}$ and training $c_{p,t}$.
Consequently, methods with high sampling costs will have less budget available for training, and will therefore afford fewer rehearsal samples.\footnote{We assume that the number of training steps needed for convergence is independent of the number of rehearsal samples, and cannot be lowered.}
In the next section, we propose a method that efficiently estimates the density of collateral damage samples, and affords the same number of rehearsal samples as random uniform rehearsal.

\section{Methodology}\label{sec:methodology}
We propose \texttt{mix-cd}, a rehearsal sampling scheme that efficiently prioritizes collateral damage samples  (see formal definition in Appendix~\ref{sec:algorithm}).
Our key idea is to estimate the collateral damage distribution at each fine-tuning iteration using only the samples mixed in during the previous iterationl.
Since these mixed-in samples are already part of the fine-tuning set, we get inference (forward pass) on them for free as part of the standard training loop. 
This makes the procedure as computationally efficient as rehearsing random uniformly.
The distribution estimation is conditioned on predefined partitions on the data distribution.
\texttt{mix-cd} allocates a higher rehearsal budget to partitions that suffer from more collateral damage.

\subsection{Collateral Damage Ratio Estimation}
The key is estimating the probability that a prior sample $z_p$ suffers from collateral damage without inferencing $z_p$ on $f'$.
We first partition the prior data distribution into $K$ bins.
At each fine-tuning iteration, we estimate the conditional probability (denoted by $\alpha_k$) that a sample from bin $k$ suffers from collateral damage.
Formally, 
\[\alpha_k := P(\phi_{f,f'}(z_p) = 1~\vert~b(z_p)=k)\]
where $b(z_p) \in [K]$ is the bin that sample $z_p$ falls in.
Once we have estimates $\hat{\alpha}_k$, we select a randomly drawn pretraining sample $z_p$ with probability $\hat{\alpha}_{b(z_p)} \cdot P(b(z_p))$.

\paragraph{Estimating $\alpha_k$.}
A straightforward scheme for estimating $\alpha_k$ is to sample uniformly from the prior distribution, perform inference on the samples using the fine-tuning model, and then compute the fraction of samples falling in bin $k$ that suffer from collateral damage.
While this provides an unbiased estimate of $\alpha_k$, it incurs additional inference costs.
To completely avoid \textit{any} additional inference, we propose estimating $\alpha_k$ at each iteration using the prior data samples mixed into the fine-tuning step during the previous iteration.
For the first iteration, the prior data samples are drawn uniformly at random with $\alpha_k$ set to $0.5$ for all $k$.
For subsequent iterations, we maintain running counts of the number of samples ($n_k$) mixed in from bin $k$, and the number of collateral damage samples ($u_k$) among them.
Specifically, at the end of each iteration, we update these counts as follows. 
Let $D_p$ be the prior data samples mixed in during the iteration.
\begin{equation}
    n_k \leftarrow n_k + \vert\{z_p \in D_p ~\vert~  b(z_p) = k\}\vert
    \label{eq:update_nk}
\end{equation}
\begin{equation}
    u_k \leftarrow u_k + \vert\{z_p \in D_p ~\vert~  b(z_p) = k, \phi_{f,f'}(z_p) = 1\}|
    \label{eq:update_uk}
\end{equation}
for all $k \in [K]$.
We then set our estimate $\hat{\alpha}_k := u_k/n_k$.
Since $D_p$ is already part of the fine-tuning set, we have the forward pass from $f'$ on them available as part of the standard training loop.
We further assume that predictions of the prior model on all prior data samples are cached, allowing us to 
compute $\phi_{f,f'}(z_p)$ at no additional inference cost.

\begin{remark}
Our estimation procedure is not (statistically) unbiased as we reuse samples seen during fine-tuning to estimate collateral damage distribution for unseen samples.
In a sense, we trade off computational cost for this bias.
Despite the bias, our scheme selects a sufficiently large number of collateral damage samples, which helps it outperform several baselines; see Section~\ref{sec:main_result_analysis}.
We also empirically confirmed the bias to be minimal in Appendix~\ref{sec:discuss-bias}.
Our conjecture is the continual exposure to new data from the fine-tuned tasks during training inflicts consistent collateral damage pattern to the prior task data distribution.
\end{remark}

\subsection{Partitioning Prior Data}
\label{sec:methodology-partition_prior_data}
The intuition behind \texttt{mix-cd} is that by partitioning the prior data distribution into bins, we can identify regions that suffer more from collateral damage.
We can then prioritize rehearsing from such regions over others during fine-tuning.
We can use any partitioning as long as the collateral damage is \textit{not} conditionally independent of the partitions.
When the collateral damage is conditionally independent of the partitions, \texttt{mix-cd} degenerates to random uniform rehearsal.

\paragraph{Desirable qualities of ``good'' partitions.}
The partition strategy is important since it directly impacts the weighted sampling. 
A good partition predicts collateral damage with high accuracy — partitioned cells either contain all or no collateral damage samples. 
Having fewer partitioned cells is another desirable quality as exploring more cells requires allocating more sampling budget to identify cells that require prioritization. 
Trading off between the two desiderata is a common challenge in machine learning. 
For example, decision trees optimize for the purity of leaf nodes but also the size of the tree to prevent overfitting.

\paragraph{Practical heuristics for partition selection.}
To avoid partitioning with ineffective bins, we calculate the KL divergence between collateral damage ratios of partitions with a uniform distribution.
A partition is effective if the KL divergence exceeds a certain threshold.
Empirically we found that 0.01 is an effective threshold for identifying effective partitions.
In practice, after the first iteration of fine-tuning with random rehearsal, the KL statistics for partitions can be calculated for partition selection with no additional computation required.
The ablation study on the selection of partitions is presented in Sec~\ref{sec:discuss_partition_strategy}.
Next, we discuss some partition strategies that work well with \texttt{mix-cd}, and are common to obtain in datasets. 

\paragraph{Partition with prior data loss.}
Prior data can be partitioned according to their losses on the prior task.
Bins can be defined based on fixed-sized loss quantiles.
Typically, examples with higher (lower) loss in prior tasks are far from the decision boundary, and thus more (less) likely to be forgotten during fine-tuning. Thus, partitioning with prior data loss is useful to identify slices where collateral damages happen more (less) frequently.

\paragraph{Partition with auxiliary information.}
Prior data can also be partitioned with auxiliary information such as class labels and/or other meta-labels.
Usually, these meta-labels convey semantic meanings that help distinguish whether certain regions would suffer more from collateral damage.
For multilingual translation datasets, the language serves as a natural partition.
For instruction-tuning datasets, the source instruction-tuning task also naturally partitions the instruction data.
In our experiments, we find that partitions based on combining prior loss and auxiliary labels perform the best.

\paragraph{Combining multiple partitions.}
Multiple partitions can be combined to form even more finer-grained partitions.
Given two partition strategies $A = {a_1, \cdots, a_n}$ and $A' = {a'_1, \cdots, a'_m}$, the combined partition is simply the set product of $A$ and $A'$ with $n \times m$ bins.
If $A$ is independent of $A'$, then the collateral damage likelihood of bin $a_i \cap a'_j$ is estimated by factoring with the individual partitions:
\[
    p(\phi|b_{a_i,a'_j}) \propto p(\phi|b_{a_i}) \cdot p(\phi|b_{a'_j})
\]
On the other hand, if $A$ is conditionally independent of $A'$ given collateral damage, then we can estimate the collateral damage likelihood by factoring and accounting for the conditional dependency:
\[
    p(\phi|b_{a_i,a'_j}) \propto \frac{p(b_{a_i})\cdot p(a'_j)}{p(b_{a_i, a'_j})} \cdot p(\phi|b_{a_i}) \cdot p(\phi|b_{a'_j})
\]
When the (conditional-)independence relation between partitions holds, estimating the collateral damage likelihood by factoring is more sample efficient since only $n + m$ statistics needed to be maintained, as opposed to $n \times m$ when estimating jointly.
In practice, we can test whether such relations hold by the end of the first iteration of fine-tuning with no additional computational cost.

\begin{figure*}
    \centering
    \includegraphics[width=\textwidth]{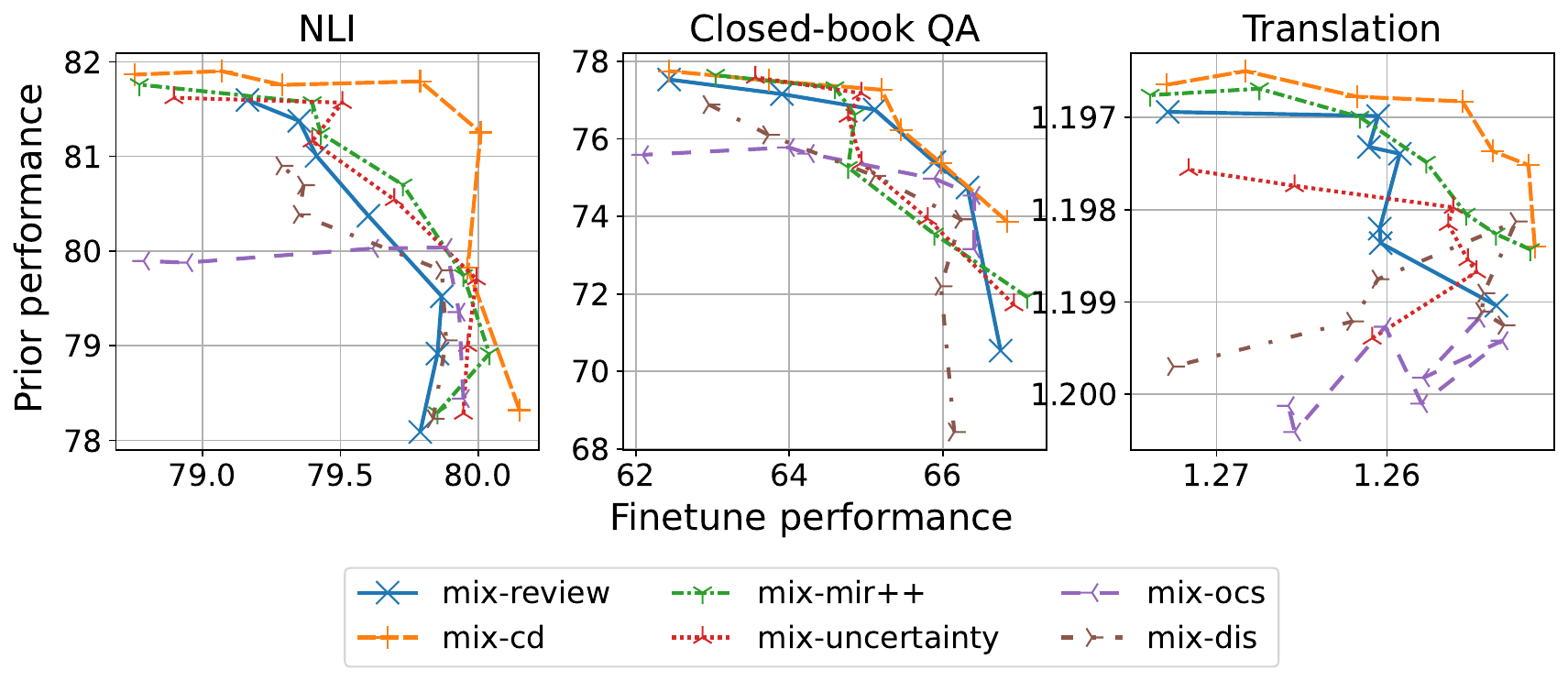}
    \caption{Pareto frontiers of prior and fine-tune performance. Curves closer to the top right are preferable.}
    \label{fig:pf-main}
\end{figure*}

\section{Experiments and Discussion}\label{sec:experiments}

\subsection{Experiment Setup}
\label{sec:experiment_setup}
We experiment on three different tasks that commonly utilize a multistage-fine-tuning pipeline: text classification, closed-book QA, and multilingual translation.
More technical details can be found in Appendix~\ref{sec:exp_technical_details}.

\paragraph{Text classification: MNLI-Scitail}
We start with a DistilBERT~\citep{sanh2020distilbert} fine-tuned on MNLI~\citep{kim2019semantic} for natural language inference (NLI), then fine-tune it on Scitail~\citep{Khot2018SciTaiLAT}, an NLI dataset for scientific statements.
The ground truth class labels and genre labels are used for partitioning.
The prior and current task performances are defined as the classification accuracy on the holdout test sets for MNLI and Scitail respectively.

\paragraph{Closed-book QA: SQuADv2-BioASQ}
We start with a RoBERTa~\citep{liu2019roberta} fine-tuned on SQuADv2~\citep{rajpurkar2018know} for general domain question answering, then fine-tune it on BioASQ~\citep{nentidis2020results}, a closed-book QA dataset for biology domain knowledge.
Binary labels of whether a sample is answerable or not are used for partitioning.
The prior and current task performances are defined as the exact-matching accuracy on the holdout datasets for SQuADv2 and BioASQ respectively.

\paragraph{Multilingual translation: translating Danish to English}
We start with \texttt{mBart50}~\citep{tang2020multilingual}, a multilingual translation model that translates from 50 different languages to English. 
We additionally fine-tuned the model on Danish, which was previously not supported by the base \texttt{mBart50} model.
The prior language labels are used for partitioning the data distribution, as we expect different languages to suffer collateral damage with different severity.
The prior task performance is defined as the average loss of all language samples excluding Danish in holdout Opus100 and the fine-tune task performance is defined as the average loss of Danish samples in holdout Opus100.

\paragraph{Training configuration}
For each experiment, we report the joint performance of the pretrain and fine-tune task on holdout datasets, evaluated at the end of fine-tuning.
The results are averaged over 5 repetitions for the NLI task, 10 for QA, and 5 for translation.
The mix ratio $\beta$ is chosen to be in the range of $[0.01, 0.9]$ such that all rehearsal methods cover similar fine-tuning performance.



\subsection{Mix-cd Outperforms Baselines}
To demonstrate the general effectiveness of \texttt{mix-cd} in diverse fine-tuning settings, we compared it with other rehearsal strategies of equal computation cost.
Recall an iteration of fine-tuning refers to fine-tuning the model on every $n$ samples.

\subsubsection{Baseline Descriptions}
Baseline methods can be classified into two categories: offline and online.
Offline methods select important prior samples to rehearse before the fine-tuning begins.
During fine-tuning, important selected samples are rehearsed randomly.
These methods are computationally efficient as they do not require additional sampling costs.
However, they suffer from lacking information regarding the new fine-tuning task since selection happens offline before fine-tuning.
Thus, the selected prior samples cannot be targeted to mitigate the incurred collateral damage.

On the other hand, online methods select samples for rehearsal when the prior samples are streamed online during fine-tuning.
Specifically, a set of $n_p$ prior samples are first randomly sampled for each batch of $n_f$ fine-tuning data.
The online method assigns a priority score to the $n_p$ prior samples and filters the top $k$~\% to mix into the batch for rehearsal.

Recall the prior rehearsal computational budget $c_p$ consists of the sampling $c_{p,s}$ and training $c_{p,t}$ cost.
The effective number of prior samples to train on depends on the sampling cost, which further depends on the cost of assigning priority scores and the filter ratio $k$.
We adopt a filter ratio of 50~\% for all online methods to balance the effectiveness selection and budget for training.
To factor in the priority assignment, we approximate the computation cost of a forward pass as half of a backward pass in terms of FLOPs.
For example, suppose the priority assignment requires one forward pass on the model.
Then the assignment is worth training 1/3 of a sample since training one sample requires one forward and one backward pass.
We calculated the effective numbers for each method (which might be different depending on the sampling cost) to control for an equivalenl total computational budget.

\paragraph{Random baseline}
\texttt{mix-review} is the name coined by \citet{he2021analyzing} for uniform-randomly   rehearsing previous task samples.
Random rehearsal is a surprisingly strong baseline in our setting where the previous task data is accessible.

\paragraph{Offline baselines} 
Online coreset selection (\texttt{mix-ocs}) is a coreset selection method proposed by \citet{yoon2022online}.
Dynamic instance selection (\texttt{mix-dis}) is a rehearsal method for continual learning proposed by \citet{mok-etal-2023-large}.
For both methods, a subset of size equivalent to the fine-tuned dataset is selected offline and rehearsed randomly during fine-tuning.

\paragraph{Online baselines}
Online methods differ in the definition of priority score.
\texttt{mix-uncertainty} prioritizes samples with high uncertainty, a common objective for active learning and data selection.
The uncertainty is estimated with prediction entropy for classification tasks and sequence log-likelihood for generative tasks.
\texttt{mix-mir++} is a modification of Maximal Interfered Retrieval (MIR)~\cite{aljundi2019online} for a computation-constraint setting. 
Typical MIR calculates the online difference in prior sample loss between the current fine-tuned model and a copy of the model with one additional gradient step on the fine-tuned data, which is too costly.
Instead, we modified their method to calculate the difference in prior sample loss between the current fine-tuned model and the cached base model. We observed the performance of \texttt{mix-mir++} to be significantly better than MIR in our Pareto frontier curves, and thus we only report the performance of \texttt{mix-mir++}.

\subsubsection{Result analysis}
\label{sec:main_result_analysis}
The main result is presented in Fig~\ref{fig:pf-main}, where \texttt{mix-cd} consistently outperforms the \texttt{random} baseline over all the experiment settings.
This supports \texttt{mix-cd} as the drop-in replacement for \texttt{random} since the performance gain comes at no additional computation cost.
Online baselines perform similar to or worse than \texttt{random} since for the given computation budget, spending the budget on sampling is not a desirable tradeoff for performance.
The performance of offline methods is the worst since the selection objective does not account for the fine-tuned task information.
This highlights the importance of the adaptivity in online methods.



\subsection{How many more collateral damage samples does \texttt{mix-cd} rehearse?}
\begin{figure}[ht]
    \centering
    \includegraphics[width=\columnwidth]{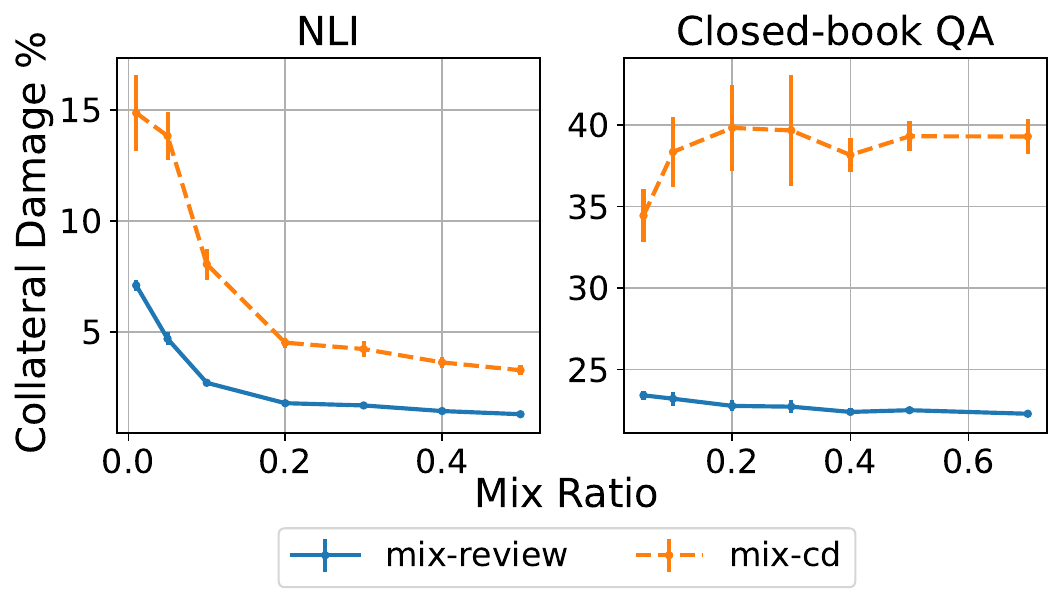}
    \caption{Proportion comparison of collateral damage per sample between random uniform and \texttt{mix-cd} across different mix ratios.
    \texttt{mix-cd} consistently samples twice or more collateral damage for rehearsal compared to random uniform, which explains the superior performance.
    }
    \label{fig:cd-proportion-ablations}
\end{figure}

The design goal of \texttt{mix-cd} is to sample collateral damage samples more efficiently.
Fig~\ref{fig:cd-proportion-ablations} compares the actual proportion of collateral damage samples in all sampled data, for \texttt{mix-cd} and random uniform sampling.
\texttt{mix-cd} consistently samples twice or more collateral damage for rehearsal compared to random uniform sampling for all mix ratios.
The empirical result supports that \texttt{mix-cd} achieves its intended purpose and also explains the superior performance over random uniform sampling.

\subsection{Selecting bins with collateral damage signal is crucial for \texttt{mix-cd}}
\label{sec:discuss_partition_strategy}
\begin{figure}[ht]
    \centering
    \begin{subfigure}{\columnwidth}
        \centering
        \includegraphics[width=\columnwidth]{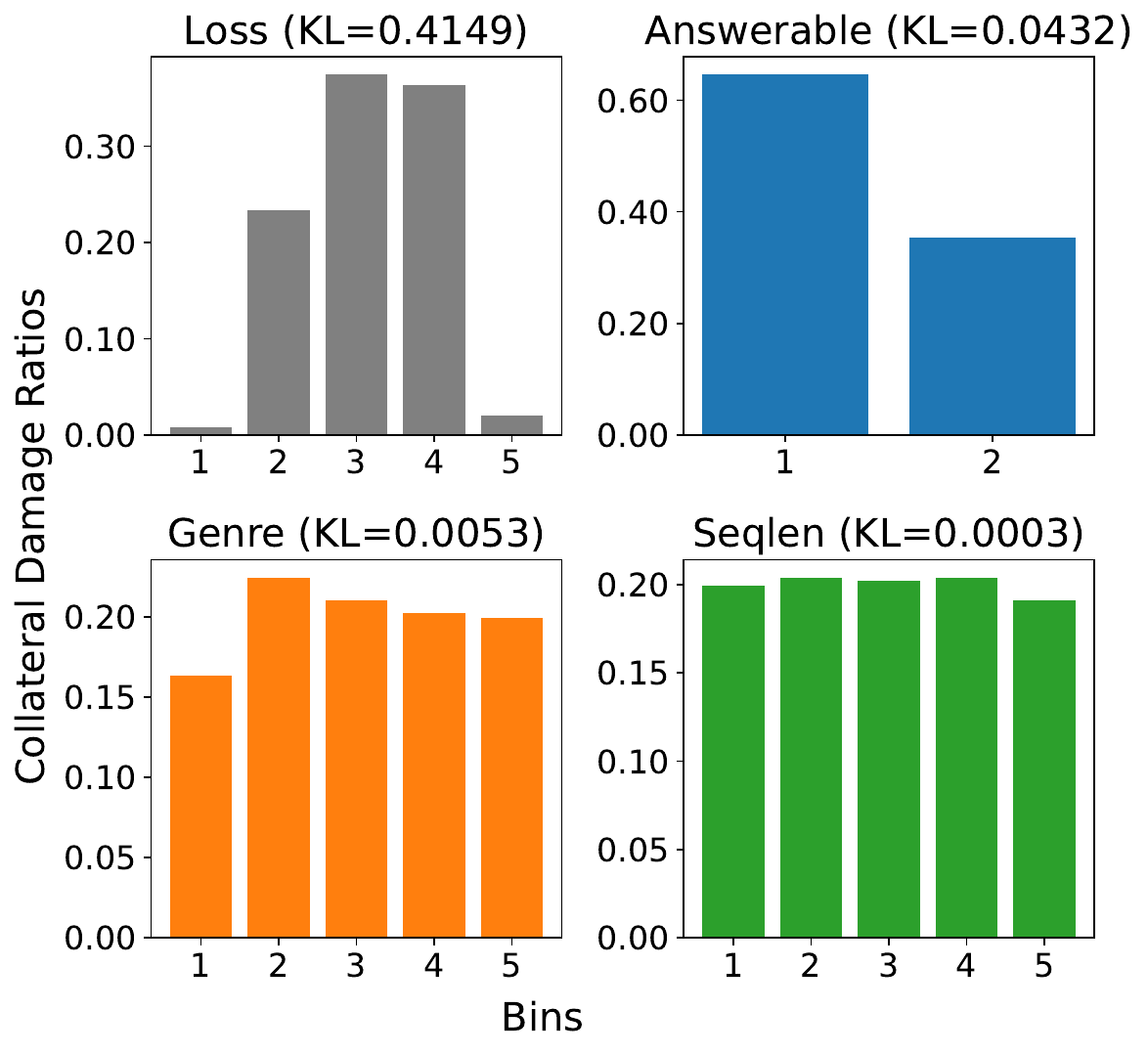}
        \subcaption{Per-bin collateral damage ratios using different partitions.}
        \label{fig:partition-ablation-bar}
    \end{subfigure}
    
    \begin{subfigure}{\columnwidth}
        \centering
        \includegraphics[width=.7\columnwidth]{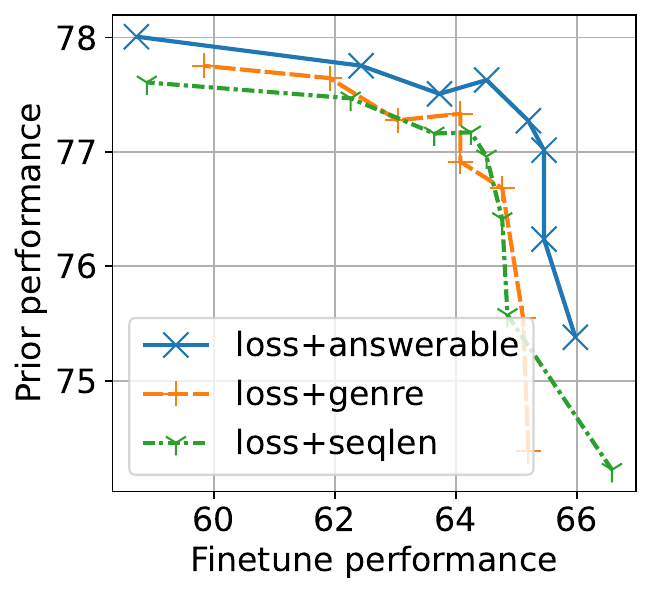}
        \subcaption{Comparing the performance of different mix-cd variants by combining loss with other partitions.}
        \label{fig:partition-ablation-curve}
    \end{subfigure}
    
    \caption{Ablation study on different partitions for the data distribution.
    Partitions with higher KL divergence in collateral damage ratios between bins (e.g. loss and answerable partitions) provide better signals for prioritizing collateral damage samples.
    }
    \label{fig:partition-ablation}
\end{figure}

\vspace{2mm}
Recall the partition selection strategy proposed in Section~\ref{sec:methodology-partition_prior_data}.
Fig~\ref{fig:partition-ablation} demonstrates the effectiveness of the selection strategy on SQuADv2.
There are four types of partitions available for SQuADv2.
Prior loss partition splits the data distribution with the prior loss values evaluated on the base model and bins them according to 5 fix-sized loss quantile intervals.
The answerable partition splits the data distribution by the binary label of whether the answer can be found in the given context.
Genre partition splits the data distribution by its genre. 
Sequence length partition splits the data distribution by the sequence length of the samples. 
After evaluating the KL divergence with the uniform distribution, the loss and answerable bins are selected as the best candidates for \texttt{mix-cd} partitions.
Fig~\ref{fig:partition-ablation-curve} verifies that coupling loss and answerable partitions is the best combination for joint performance.

\subsection{What if meta-information for partitions is not included in the dataset?}
\begin{figure}[ht]
    \centering
    \includegraphics[width=\columnwidth]{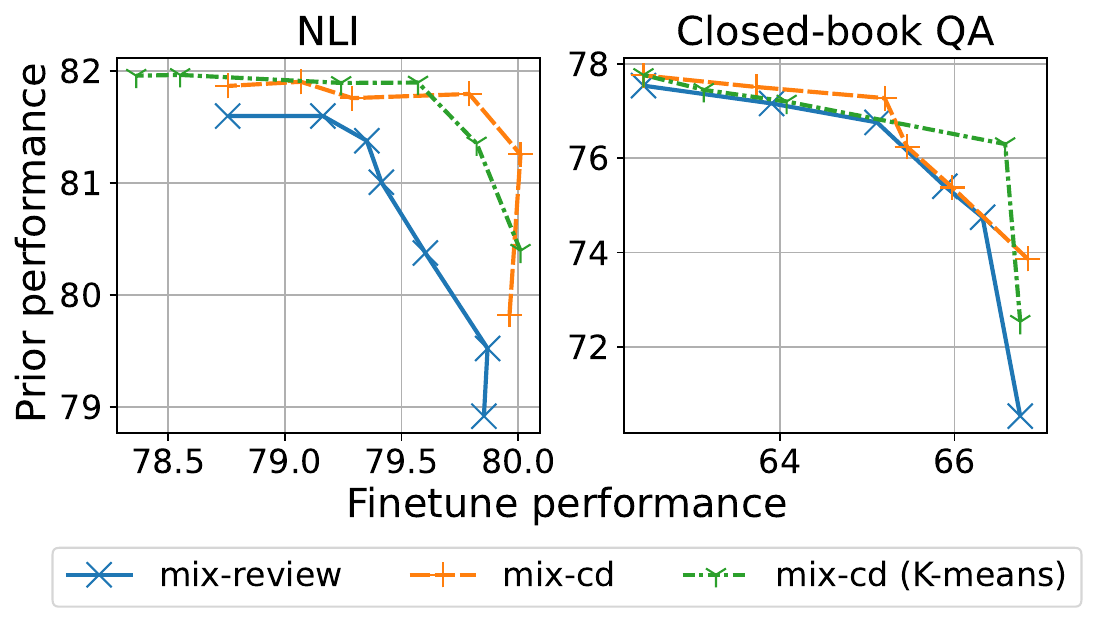}
    \caption{Ablation study on unsupervised clustering partitions with embeddings generated with the prior base model.
    The general partition can serve as a good alternative when meta-data for partitioning is unavailable.
    }
    \label{fig:unsupervised-ablations}
\end{figure}

In previous experiments, we demonstrated that meta-information in the dataset (e.g. supervised labels, categorization) can serve as effective partitions.
In real-world applications, the dataset may lack any type of natural partitioning.
In these cases, partitions generated from unsupervised techniques can serve as alternatives to provide effective signals for identifying collateral damage.
Here we demonstrate the effectiveness of the most general unsupervised setting: K-means clustering with latent embeddings generated by the base model.
Fig~\ref{fig:unsupervised-ablations} shows the effectiveness of K-means clustering matches that of the natural partitions.
The semantic similarity encoded in the embeddings likely also contains information for predicting the likelihood of collateral damage.
We conclude that partitions generated with unsupervised clustering can be equally effective, which further enhances the general applicability of \texttt{mix-cd} as a rehearsal strategy.

\subsection{What if the prior task performance signal is not readily available?}
\texttt{mix-cd} relies on the assumption that prior task performance signal in the form of predictions or losses is cached and available for free.
Even though saving the information during previous training is easy and essentially free, it is possible for the assumption to not hold, likely because the model was trained by another party.
For these scenarios, the most straightforward solution is to perform inference and cache the model information for later use.
Fortunately, the inference computation cost for inferencing can be amortized over future fine-tuning of the model.
For example, the DistilBERT model trained on MNLI (\href{https://huggingface.co/typeform/distilbert-base-uncased-mnli}{\texttt{typeform/distilbert-base-uncased-mnli}}) in our experiments was downloaded 50k times last month (Jan 2025) on HuggingFace, easily justifying the one-time inference cost after amortization.

\section{Conclusion}
We proposed a rehearsal-based sampling strategy to prioritize collateral damage samples during fine-tuning.
The simplicity and effectiveness make it an appealing drop-in replacement for the typical random uniform rehearsal strategy.
Future work can investigate better hybrid methods combining both rehearsal and weight regularization for forgetting prevention.

\paragraph{Limitations}
We assume the last-epoch prediction or loss of the prior data on the base model is saved during the fine-tuning phase.
The loss or prediction information provides important signals to identify collateral damage regions in the prior data distribution.
More investigation is also needed to examine whether the original prior performance can be fully recovered with \texttt{mix-cd}.
Non-uniform rehearsal with \texttt{mix-cd} may prioritize the region suffering from the most collateral damage.
This might introduce bias in the fine-tuned model that cannot be detected merely with the prior task performance.
Further study is required to examine whether collateral damage in minority sample regions is affected by the rehearsal scheme.

\bibliography{refs}

\appendix

\section{Experiment Technical Details}
\label{sec:exp_technical_details}

\subsection{Text classification: MNLI-Scitail}
We first fine-tune a DistilBERT model on MNLI~\citep{kim2019semantic}, which is a natural language inference (NLI) dataset, and then the model fine-tune on Scitail~\citep{Khot2018SciTaiLAT}, a natural language entailment dataset for scientific statements.
NLI tasks aim to determine the relationship (entailment, contradiction, or neutral) between a pair of input sentences.
The model is fine-tuned with AdamW with learning rate of $2\cdot 10^{-6}$ and weight decay of $10^{-5}$.
There are 393,000 samples in the MNLI pretrain training dataset.
In addition to relation labels, additional genre labels (e.g. fiction, government, travel) for the sentence pairs are also provided.
To implement \texttt{mix-cd}, we use the ground truth class labels and genre labels for partitioning.
For each iteration, we fine-tune with 1,000 samples from the Scitail training set (iterating over the entire training set of 23,600 samples after 25 iterations).
The pretrain and fine-tune task performances are defined as the classification accuracies on MNLI and Scitail, respectively.

\subsection{Closed-book QA: SQuADv2-BioASQ}
We first fine-tuned a RoBERTa~\citep{liu2019roberta} on SQuADv2~\citep{rajpurkar2018know} for general domain closed-book QA, then fine-tuned it on BioASQ~\citep{nentidis2020results}, a closed-book QA dataset for biology domain knowledge.
The model is fine-tuned with AdamW with a learning rate of $1\cdot 10^{-5}$, warming up the learning rates from $1\cdot 10^{-7}$ for 5 iterations, then cosine annealing the learning rate to $1\cdot 10^{-6}$, and weight decay of $10^{-5}$.
There are 130K samples in the SQuADv2 training dataset.
To implement \texttt{mix-cd}, we use the binary labels of whether a sample is answerable or not are used for partitioning.
For each iteration, we fine-tune with 1,000 samples from the BioASQ training set for 20 iterations.
The prior and current task performances are defined as the exact-matching accuracy on the holdout datasets.

\subsection{Multilingual translation: translating Danish to English}
The experimental setting for multilingual translation is slightly different from classification tasks.
Instead of fine-tuning on a new dataset, we take a multilingual translation model that translates from 50 different languages to English and fine-tune it to perform translation on one additional language.
To implement \texttt{mix-cd}, we use the language type for partitioning.
We would like to prevent any deterioration in the performance of the existing 50 languages due to fine-tuning.
It is expected for the translation for some languages in the pretrain language to deteriorate after fine-tuning.
We leverage the pretrain language as auxiliary information for partitioning to identify and fix the languages with more collateral damage.

The base model of choice is \texttt{mBart50}~\citep{tang2020multilingual}, a generative language model pretrained on translation sentence pairs of 50 different languages to English.
The model is fine-tuned with AdamW with learning rate of $10^{-5}$ and weight decay of $10^{-5}$.
The training data pairs (both prior and fine-tuned) are taken from Opus100, a multilingual, English-centric dataset that consists of sentence pairs translating from 100 other languages to English.
We fine-tune the model on Danish, which was previously not supported by the pretrained \texttt{mBart50} model.
For each iteration, we subsample 10,000 new Danish-English sentence pairs to fine-tune.
The prior dataset consists of 10,000 random uniform samples from the languages that \texttt{mBart50} was originally capable of translating. 
The prior task performance is defined as the average loss of all prior language samples and the fine-tune task performance is defined as the average loss of Danish samples.

\section{Ablation Studies}

\subsection{\texttt{mix-cd} reduces forgetting by equalizing the performance of different bins}

\begin{figure}[ht]
    \centering
    \begin{subfigure}{\columnwidth}
        \centering
        \includegraphics[width=\columnwidth]{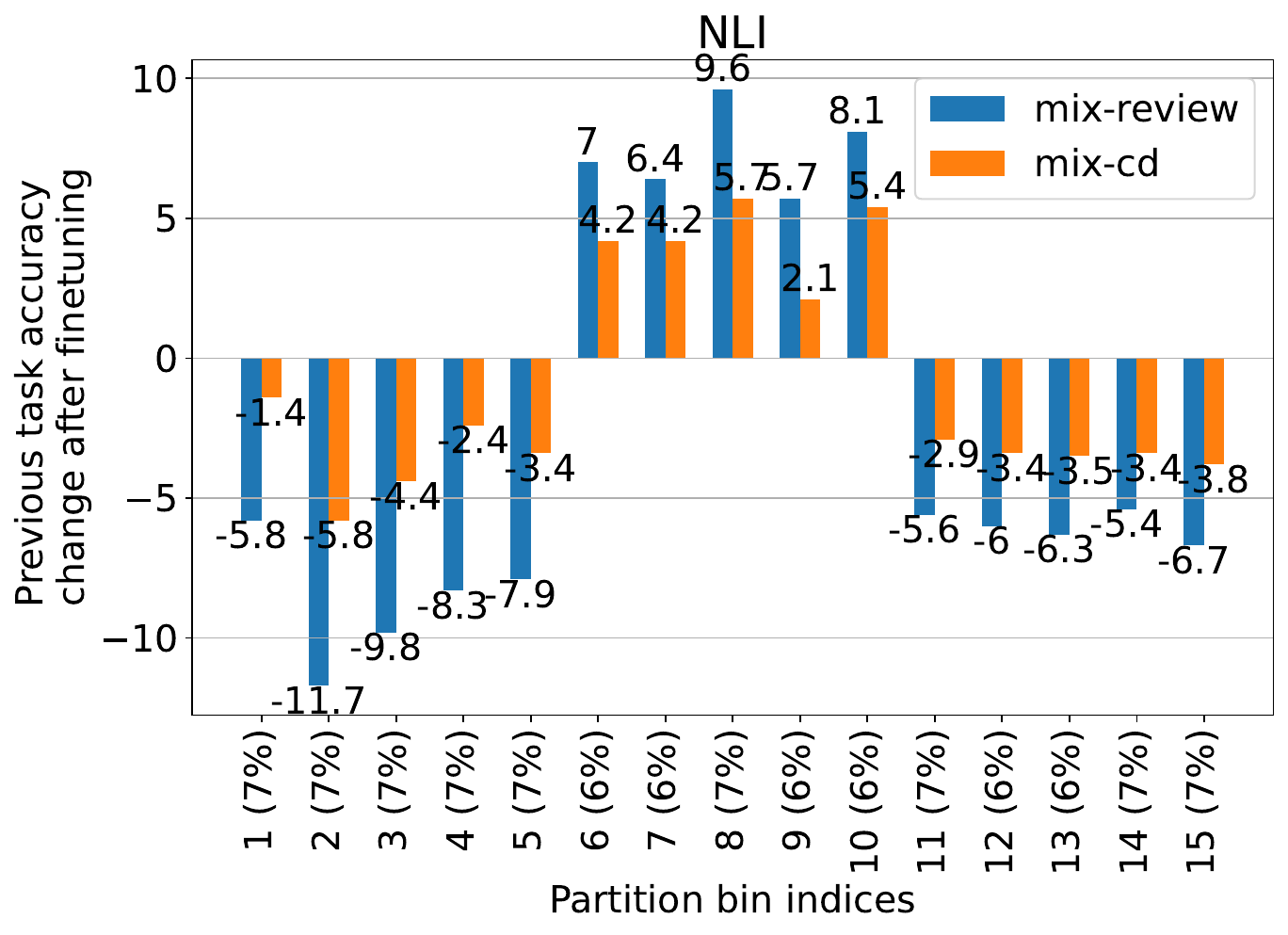}
    \end{subfigure}
    
    \begin{subfigure}{\columnwidth}
        \centering
        \includegraphics[width=.9\columnwidth]{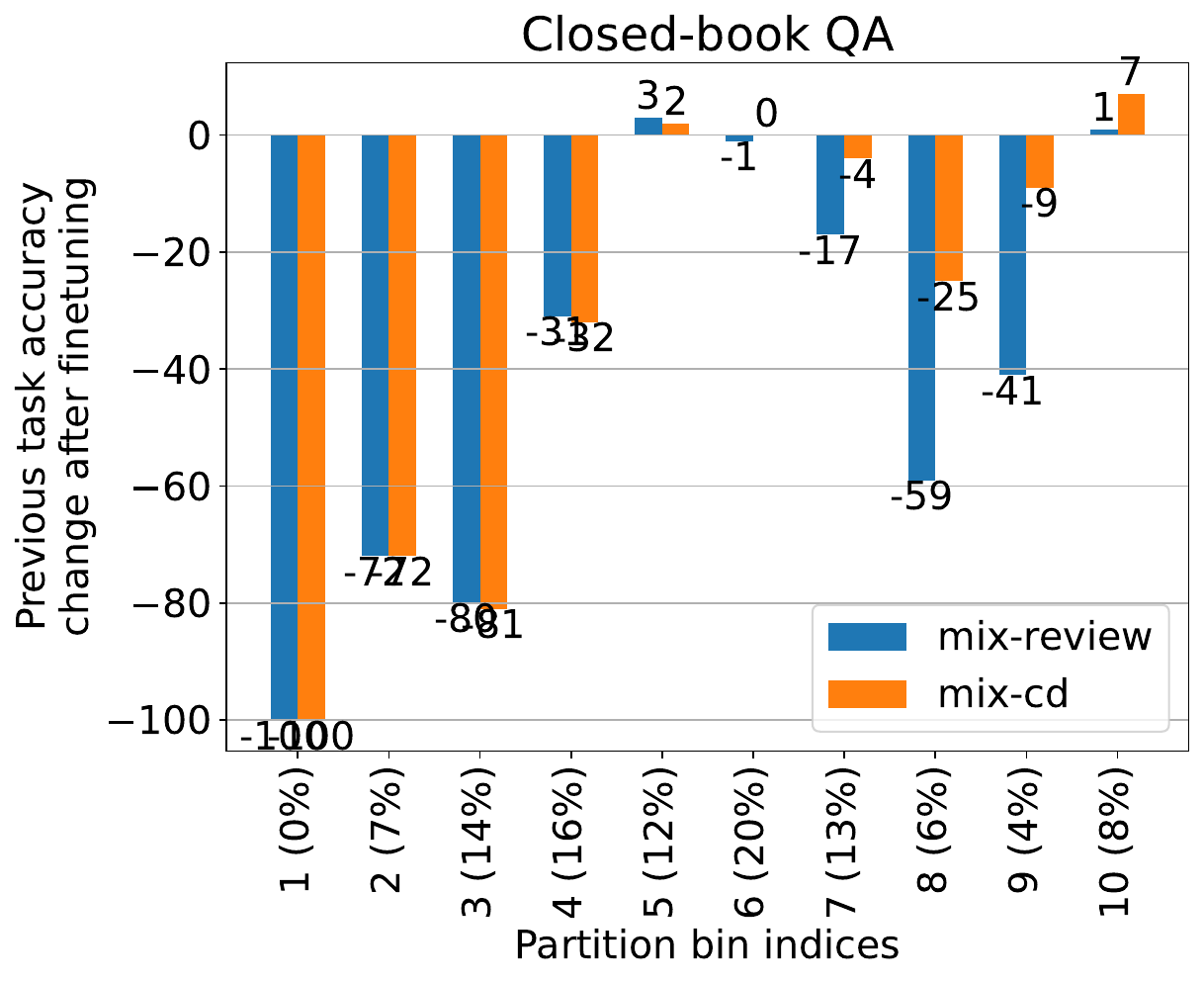}
    \end{subfigure}
    \caption{
    Per-bin performance difference before and after fine-tuning.
    Compared to random uniform rehearsal, \texttt{mix-cd} achieves a more balanced performance amongst bins, utilizing the rehearsal budget more efficiently, thereby achieving a more desirable tradeoff between the performance of the new and prior tasks.
    }
    \label{fig:per-bin-performance}
\end{figure}

Our understanding of why \texttt{mix-cd} performs better is through non-uniformly rehearsing the previous task data distribution, the computation budget can be better spent on fixing collateral damages caused by the new finetuning task. 
We presented Figure~\ref{fig:cd-proportion-ablations} to show that \texttt{mix-cd} consistently samples more collateral damage samples compared to random uniform.
To make the argument complete, we will explicitly draw the connection between ``sampling more collateral damages'' and ``better retaining the previous task performance''. 
We present the performance change (relative to the base model) of each partition and compare random uniform versus \texttt{mix-cd} rehearsal in Figure~\ref{fig:per-bin-performance}. 
We observe that \texttt{mix-cd} achieves better overall performance by allocating the budget to fix the most forgotten partitions.
We highlight that since collateral damage does not happen uniformly in the previous task data distribution, rehearsing non-uniformly (and targeting the more impacted partitions) is the key to maintaining uniform performance across partitions.

\subsection{Combining multiple partition strategies benefits \texttt{mix-cd}}
\label{sec:discuss-partition-combination}
\begin{figure}[ht]
    \centering
    \includegraphics[width=\columnwidth]{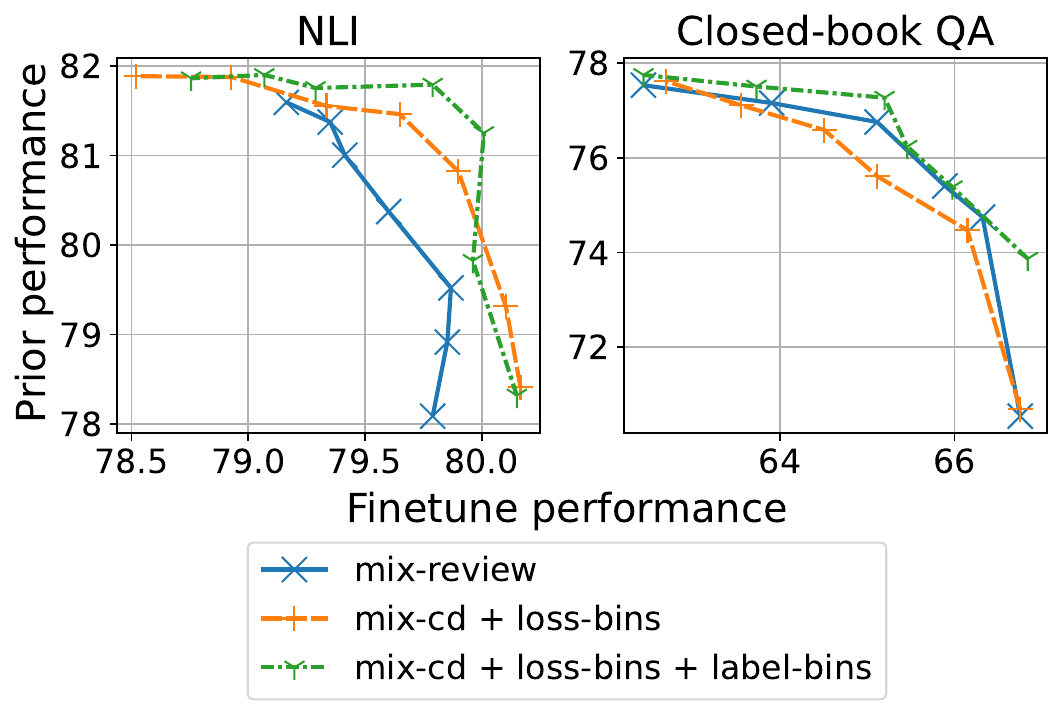}
    \caption{
    Ablating \texttt{mix-cd} by comparing single partition strategy with combining multiple partitions.
    Combining partitions is better at characterizing high-collateral-damage regions in the data distribution.
    This benefits the prior task performance, especially in low-mix-ratio settings where budget allocation is more crucial.
    }
    \label{fig:pc-bin-ablation}
\end{figure}

We introduced different partitioning strategies in Section~\ref{sec:methodology-partition_prior_data} and recommended combining multiple partitions to maximize the modeling of collateral damage distribution.
In Figure~\ref{fig:pc-bin-ablation} we compared single partition versus multiple partitions and found that the latter consistently outperforms, especially in settings where the budget allocated to rehearsal is scarce (right part of the figure).
Thus, we empirically confirmed that combining more diverse partitions is more beneficial to the prediction of collateral damage and the accurate prediction translates into more effective rehearsal under equal computation budgets.

\subsection{Confirming the bias in collateral damage estimation for \texttt{mix-cd} is insignificant}
\label{sec:discuss-bias}
\begin{figure}[ht]
    \centering
    \includegraphics[width=\columnwidth]{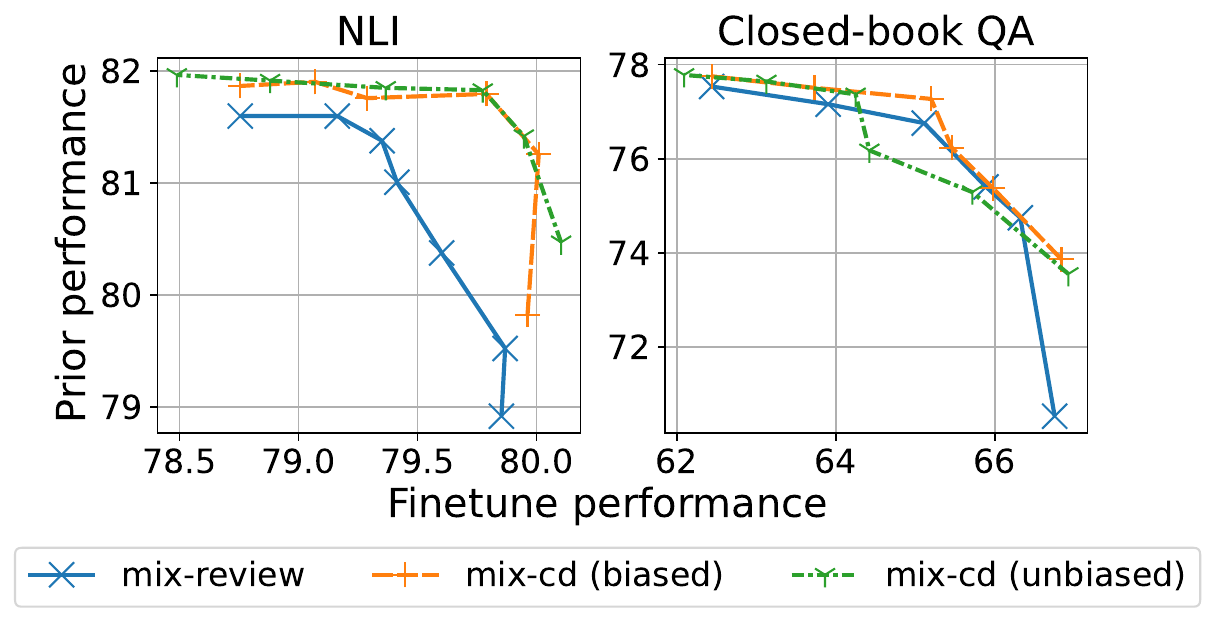}
    \caption{
    Comparing the Pareto curves of biased and unbiased versions of \texttt{mix-cd}.
    The results suggest the statistical bias of \texttt{mix-cd} does not compromise the collateral damage ratio estimation and achieves comparable performance as the unbiased, but computationally-expensive counterpart.
    }
    \label{fig:pc-bias-ablation}
\end{figure}

We explore the impact of the statistically biased estimation of collateral damage ratios when reusing samples that were already trained on. 
Reusing the samples allows no extra computation at the cost of biased ratio estimation.
We compared the \texttt{mix-cd} with the unbiased version.
For the unbiased version of mix-cd, we estimate the collateral damage ratios by inferencing a holdout set of examples (requiring additional computation for forward passes). 
The results are presented in Figure~\ref{fig:pc-bias-ablation}.
We found that the theoretically biased and unbiased versions of mix-cd perform similarly in practice, indicating that reusing the forward pass information from trained samples is a good proxy. 
We suspect the bias is negligible because the model is continuously being exposed to more new finetuned task signals, which will continue to cause collateral damage to the partitions. 
Therefore, rehearsing a sample from a specific partition would likely not fix the collateral damage of the partition once and for all.

\section{Algorithm of \texttt{mix-cd}}
\label{sec:algorithm}
\begin{algorithm}[ht]
   \caption{\texttt{mix-cd}}
   \label{alg:mix_cd_sample}
    \begin{algorithmic}[1]
        \STATE {\bfseries Input:} number of iterations $N$, prior dataset $Z_p$, fine-tune dataset $Z_f$, base model $f$, mix ratio $\beta$, number of partitions $K$, number of training samples per iteration $n$.
        \STATE // Initialize estimates $\hat{\alpha}_k$
        \FOR{$k=1$ {\bfseries to} $K$}
            \STATE Initialize $\hat{\alpha}_k \leftarrow 0.5$
            \STATE Initialize $u_k \leftarrow 0, n_k \leftarrow 0$
        \ENDFOR
        \STATE Initialize fine-tune model $f' \leftarrow f$
        \FOR{$n=1$ {\bfseries to} $N$}
            \STATE Initialize dataset $D_f \leftarrow$ $(1 - \beta) \cdot n$ random uniform samples from $Z_f$
            \STATE Initialize dataset $D_p \leftarrow \{\}$
            \REPEAT
                \STATE $z_p \leftarrow$ sample from $Z_p$ with probability $\hat{\alpha}_{b(z_p)}\cdot P(b(z_p))$
                \STATE $D_p \leftarrow D_p \cup \{z_p\}$
            \UNTIL{$\vert D_p\vert \geq \beta \cdot n$}
            \STATE Train $f'$ for one iteration on $D_f~\cup~D_p$
            \STATE // Update estimates $\hat{\alpha_k}$
            \FOR{$k=1$ {\bfseries to} $K$}
                \STATE Update $u_k, n_k$ according to Eq~\ref{eq:update_nk} and~\ref{eq:update_uk}
                \STATE $\hat{\alpha}_k \leftarrow u_k/n_k$
            \ENDFOR
        \ENDFOR
    \end{algorithmic}
\end{algorithm}

\end{document}